\documentclass[11pt]{article}
\usepackage[margin=1in,headheight=13.6pt]{geometry}
\usepackage{hyperref}
\usepackage[parfill]{parskip}
\usepackage{setspace}
\usepackage{libertine}
\usepackage{graphicx}
\usepackage{fancyhdr}
\usepackage{comment}
\usepackage{amsmath}
\usepackage{amsfonts}
\usepackage{listings}
\usepackage{color}
\usepackage{xcolor}

\definecolor{codegreen}{rgb}{0,0.6,0}
\definecolor{codegray}{rgb}{0.5,0.5,0.5}
\definecolor{codepurple}{rgb}{0.58,0,0.82}
\definecolor{backcolour}{rgb}{0.95,0.95,0.92}
 
\lstdefinestyle{mystyle}{
    backgroundcolor=\color{backcolour},   
    commentstyle=\color{codegreen},
    keywordstyle=\color{magenta},
    numberstyle=\tiny\color{codegray},
    stringstyle=\color{codepurple},
    basicstyle=\footnotesize,
    breakatwhitespace=false,         
    breaklines=true,                 
    captionpos=b,                    
    keepspaces=true,                 
    numbers=left,                    
    numbersep=5pt,                  
    showspaces=false,                
    showstringspaces=false,
    showtabs=false,                  
    tabsize=2
}
\usepackage[
    backend=bibtex,
    style=nature,
    sorting=none]{biblatex}
    
\def\keywordname{{\bfseries \emph Keywords}}%
\def\keywords#1{\par\addvspace\medskipamount{\rightskip=0pt plus1cm
\def\and{\ifhmode\unskip\nobreak\fi\ $\cdot$
}\noindent\keywordname\enspace\ignorespaces#1\par}}

\title{Spatial Analysis Made Easy with Linear Regression and Kernels}

\author{
  Philip Milton\\
  \small{MRC Centre for Outbreak Analysis and Modelling}\\
  \small{Department of Infectious Disease Epidemiology}\\
  \small{Imperial College London, London, UK}\\
  \small{\texttt{PM5215@ic.ac.uk}}
  \and
  Emanuele Giorgi\\
  \small{CHICAS, Lancaster Medical School}\\
  \small{Lancaster University, Lancaster, UK}\\
  \small{\texttt{e.giorgi@lancaster.ac.uk}}
  \and
  Samir Bhatt\\
   \small{MRC Centre for Outbreak Analysis and Modelling}\\
  \small{Department of Infectious Disease Epidemiology}\\
  \small{Imperial College London, London, UK}\\
  \small{\texttt{s.bhatt@ic.ac.uk}}
}
\date{\today}

\begin{document}

\maketitle

\begin{abstract}

Kernel methods are an incredibly popular technique for extending linear models to non-linear problems via a mapping to an implicit, high-dimensional feature space. While kernel methods are computationally cheaper than an explicit feature mapping, they are still subject to cubic cost on the number of points. Given only a few thousand locations, this computational cost rapidly outstrips the currently available computational power. This paper aims to provide an overview of kernel methods from first-principals (with a focus on ridge regression), before progressing to a review of random Fourier features (RFF), a set of methods that enable the scaling of kernel methods to big datasets. At each stage, the associated R code is provided. We begin by illustrating how the dual representation of ridge regression relies solely on inner products and permits the use of kernels to map the data into high-dimensional spaces. We progress to RFFs, showing how only a few lines of code provides a significant computational speed-up for a negligible cost to accuracy.  We provide an example of the implementation of RFFs on a simulated spatial data set to illustrate these properties. Lastly, we summarise the main issues with RFFs and highlight some of the advanced techniques aimed at alleviating them.  

\end{abstract}

\section*{Introduction}
Spatial analysis can be seen as a supervised learning problem where we seek to learn an underlying function that maps a set of inputs to an output of interest based on known input-output pairs. In supervised learning, it is crucial that the mapping is done such that the underlying function can accurately predict (or generalise) to new data.  Generally, the output, or \textit{response variable}, is a variable measured at multiple geolocated points in space (e.g. case counts for a disease under investigation \cite{gething2016mapping},  anthropocentric indicators like height and weight \cite{osgood2018mapping, josepha2019understanding}, or socioeconomic indicators such as access to water or education \cite{graetz2018mapping, andres2018geo}). Here, we shall consider response variables as $\textbf{y} = (y_{1}, y_{1},...,y_{N}) \in \displaystyle \mathbb{R}^N$, indicating that the metric of interest is a vector of length corresponding to the $N$ locations. The inputs are referred to as \textit{explanatory variables} or \textit{covariates} and consist of multiple independent variables taken at the same $N$ locations as the response variables. Epidemiological examples of explanatory variables are population, age, precipitation, urbanicity and spatial or space-time coordinates. Explanatory variables are given as $X = [\mathbf{x}_{1},\mathbf{x}_{2},...,\mathbf{x}_{N}] \in \displaystyle \mathbb{R}^{N\times d}$, i.e. a matrix (signified by the upper case $X$) of length (rows) corresponding to the $N$ locations and width (columns) of $d$ representing the number of explanatory variables at each location. This matrix is often referred to as the \textit{design matrix} or \textit{basis}, with each row of the matrix represents a location and each column an explanatory variable. The final step is to define a measurement equation, $\textbf{y} = f\!(X) + \epsilon$, that links our explanatory variables to our responses.  


The goal of this paper is to cast complex spatial methods into a tool familiar to most researches - the linear model. We will introduce a large body of theory known as model-based geostatistics \cite{diggle1998model} from a machine learning perspective with the goal of slowly build the theory from scratch and arrive at a predictive algorithm capable of making inferences. To arrive at this goal, we will first introduce linear regression and a more powerful variant of linear regression called ridge regression. We will then introduce kernels as a way to create complex functions and show how kernels can be learnt in the linear regression framework.  Finally, we introduce a powerful new approach, random Fourier features, that is computationally favourable. We present code wherever relevant, include a brief toy example in R where we fit a spatial problem using nothing more than linear regression and some transforms. 

Within this paper, we focus on real-valued response variables as this scenario is easier to handle computationally and also better at illustrating the mathematical theory of Gaussian processes. There is considerable overlap between our introduction of kernel learning and the more traditional formulations based on Gaussian process regression. For an introduction to the Gaussian process and model-based geostatistics, we refer the reader here \cite{Rasmussen2005, Diggle2007}. For a detailed description of the mathematical correspondence between kernels and Gaussian processes, we refer the reader here \cite{kanagawa2018gaussian}.  

\section*{Linear Regression}
One of the most popular choice of model in supervised learning is the linear model. A linear model assumes that the outputs are a weighted linear combination of the inputs with some additional uncorrelated (independent) noise, $\epsilon_{i}$, \cite{McCullagh:1989} such that for any location 
\begin{equation}
\begin{split}
  y_{i} & = x_{i,1}w_{1} + x_{i,2}w_{2} + ... + x_{i,d}w_{d} + \epsilon_{i} \\
  y_{i} & = \sum_{j=1}^{d} x_{i,j}w_{j}+ \epsilon_{i} \; \; \; ,\textup{for}\;  j = 1,2,...,d
\end{split}
\end{equation} 
Where $\textbf{w} = (w_{1},w_{2},...,w_{d})\in \displaystyle \mathbb{R}^d$ represents the column vector of weights (or coefficients) of length $d$ used to transform the input to the outputs. We can write this much more succinctly using matrix notation: 
\begin{equation}    
    \mathbf{y} = X\mathbf{w}+ \mathbf{\epsilon}
\end{equation}

By changing the values of the weights, $\mathbf{w}$, we can generate a wide range of different functions. The learning task is to find an optimal set of weights (or more broadly parameters) that can produce a function which reproduces our data as close as possible. To do this, we first need to define what is meant by  "as close as possible". Mathematically, we need to define a function, known as the object function, that measures the quality of our model given a set of weights. In the case of linear regression, a common objective function is given by:
\begin{equation}
S(\mathbf{w}) = \frac{1}{2} \left \|\boldsymbol{ y }- X\mathbf{w} \right \|^2 
\end{equation}

This objective function is often called the squared loss function and computes the sum of the squared differences between the measured response variables and the model's prediction of the responses for a given set of weights. Note, the multiplication by half only added to make taking the derivative simpler and does not affect the solutions.  

The learning task is therefore to find the set of weights that minimise (or maximise the negative of) the objective function and correspondingly produce the best possible solution for our model. Formally we seek to solve: $\min_{\mathbf{w }\in \mathbb{R}^d} \left \|\boldsymbol{ y }-X\mathbf{w} \right \|^2$

Elementary calculus tells us that a minima of a given function can be found when its derivative with respect to the parameters are zero. i.e. 
\begin{equation}
    \frac{\partial S(\mathbf{w})}{\partial \mathbf{w}} = X^{T}(X\mathbf{w} - \mathbf{y})=0
\end{equation}
Rearranging the derivative allows us to solve for the optimal set of weights, $\textbf{w}^{*}$, that minimise the objective function
\begin{equation}
\textbf{w}^{*} = (X^{T}X)^{-1}X^{T}\textbf{y}
\end{equation}

Equation 5 is termed the ordinary least squares (OLS) solution. With these optimal weights, the model can predict the response for any given set of inputs. For a new unseen data point, $\textbf{x}_{*}$, the model can make a prediction of the response, $\widehat{y}_{*}$, computed by $\widehat{y}_{*} = \textbf{x}_{*}\textbf{w}^{*}$

Minimising the objective function to find weights that best fit the data is often referred to as training. Once training is done, we often need to check if predictions from this model can generalise well to new data not use in training, known as the testing. A good model should be capable of both fitting the training data well but also able to predict well to new unobserved locations.

Linear regression is ubiquitous across all fields of science due to the ease of derivation, the simplicity of the solutions,  and the guarantee that when the assumptions of the Gauss-Markov theorem are met the OLS estimator is a best linear unbiased estimator (BLUE) (see \cite{davidson2004econometric} for the standard proof). However, this does not mean that linear regression will always yield good results. Some problems are non-linear such that even the best linear model is inappropriate (we will address this using kernels in the following sections). In other scenarios, the data often violate the assumptions that guarantee OLS as the BLUE.

One of the most common violated assumptions is that the explanatory variables are uncorrelated (or orthogonal). In the most extreme case, one explanatory variable can be an exact linear combination of the one or multiple other variables, termed perfect multicollinearity \cite{farrar1967multicollinearity}. A common cause for perfect multicollinearity is in overdetermined systems when the number of explanatory variables exceeds the number of data points (the width of $X$ is greater than its length, $d > N$),  even though there may be no relationship between the variables. Both multicollinear and overdetemined systems are common in epidemiological and genetic studies \cite{vatcheva2016,yoo2014study}. When the data contains perfect multicollinearity, the matrix inversion $(X^{T}X)^{-1}$ is no longer possible, preventing the solving for the weights in Equation 5. Multicollinearity (as well as other violations to the Gauss-Markov assumptions) is characterised by a poor performance in model testing and aberrant weights often despite good performance in training.

\section*{The Bias-Variance Trade-off}
Surprisingly, one of the biggest problems with OLS is that it is unbiased. When there is a large number of predictors (i.e. $d$ is large), having an unbiased estimator can greatly overfit the data, resulting in very poor predictive capability. A celebrated result in machine learning, the bias-variance decomposition \cite{Geman1992, Domingos2000} explains why this is the case for the squared loss function. For the squared loss, and a linear model $X\mathbf{w}$, some data $y$ and the true function $f(X)$ the loss can be decomposed as a sum of expectations as:

\begin{equation}
    \begin{split}
         \mathbb{E}[(y-X\mathbf{w})^2] &= \mathbb{E}[X\mathbf{w}-f(X)]^2 + \left (\mathbb{E}[X\mathbf{w}^2]-( \mathbb{E}[X\mathbf{w}])^2  \right ) \\
         loss &= bias^2 + variance
    \end{split}
\end{equation}

Here, the bias term can be thought of as the error caused by the simplifying assumptions of the model. The variance term tells us how much the model moves around the mean. The more complex a given model, the closer it will be able to fit the data and the lower its bias.  However, a more complex model can "move" more to capture the data points resulting in a larger variance.  Conversely, if the model is very simple it will have low variance but might not be able to fit a complex function resulting in high bias. This creates a trade-off, reducing bias means increasing variance and vice versa. Figure 1 allows us to visualise the consequences of this trade-off (Supplementary Code 1). 

\begin{figure}[ht]
  \includegraphics[width=\textwidth,height=\textheight,keepaspectratio]{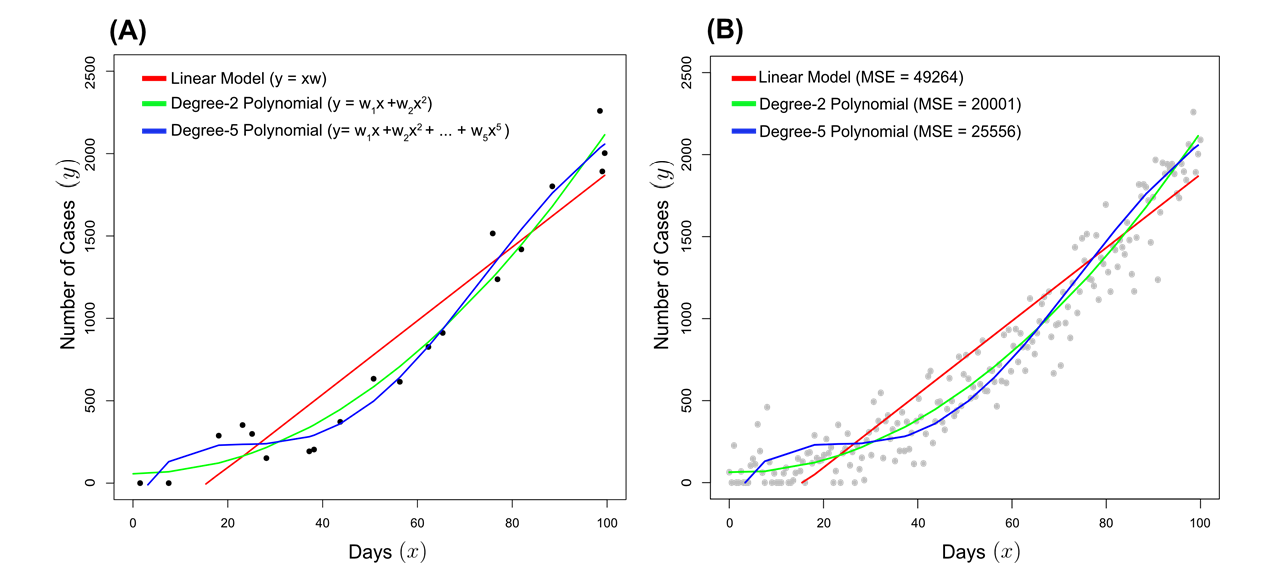}
  \caption{An example of bias-variance trade-off for a linear/degree-1 (red), degree-2 (green) and degree-5 (blue) polynomial model given the same training data. (\textbf{A}) All three models were fitted to the same subset of 20 points derived a simulated epidemic with a latent function $f(X)=1+X+0.2X^{2}$ with added Gaussian noise $(\epsilon \sim N(\mu=0,\sigma^{2}=150))$. The function is left truncated such that the number of cases is always greater than zero.  (\textbf{B}) The mean squared error (MSE) for each model was calculated for the remaining data not used for the fitting, so-called testing data.}
  \label{fig.1}
\end{figure}

In Figure 1, data points were generated from a simulated epidemic, with the y-axis representing the number of cases and the x-axis time measured in days. The true latent function of the simulated epidemic is a degree-2 polynomial with added Gaussian noise given by $y=1+X+0.2X^{2}+\epsilon$, where $\epsilon \sim N(\mu=0,\sigma^{2}=150)$. The function is left truncated such that the number of cases is always greater than zero. Three Polynomial models of different degrees were fitted (linear/degree-1 in \textit{red}, degree-2 in \textit{green} and degree-5 in \textit{blue}). The most complex, degree-5 polynomial has the closest fits to the training points in Figure 1A. However, when the error is calculated for the testing data (not used to train the model), the degree-2 model has the lowest error, shown in Figure 1B. 


The crucial point here is despite the degree-5 model fitting the training data better than the degree-2 model, it poorly predicts data not included in the training data-set. The extra parameters in the degree-5 model allow the model to move to meet all of the training points resulting in a very low bias. However, this movement to meet the training points results in very high variance, with the model fitting to the random noise in the data. Therefore, the degree-5 polynomial can be said to overfit the data, with the model's high variance out-weighting its low bias. The opposite is true of the linear model, where the linear functions cannot capture the non-linearity in the data resulting in a high bias that outweighs its low variance. The linear model underfits the data. The degree-2 polynomial strikes a balance between bias and variance, with the functions flexible enough to capture the non-linearity of latent function without fitting to the noise, capturing an optimal balance between bias and variance. 

An approach that is frequently adopted to maintain optimal bias-variance trade-off is to restrict the choice of functions in some way. Such a restriction is referred to as regularisation and can constrain weights values, stabilise matrix inversion, and prevent over-fitting.

\section*{Ridge Regression}
The most common choice of regularisation is called Tikhonov regularisation or ridge regression/weight decay in the statistical literature \cite{Tikhonov1963, Bell1978, Hoerl1970}. The fundamental idea behind ridge regression is to prevent overly complex functions by favouring weights and thus functions with small norms (small absolute values). This is achieved by adding a penalisation term with respect to the norm of the weights to our objective function giving us the standard objective function for ridge regression: 
\begin{equation}
    S(\mathbf{w}) = \frac{1}{2}(\textbf{y}-X\mathbf{w})^2 + \frac{1}{2}\lambda \left\|\mathbf{w}\right \|^2
\end{equation}
The objective function for ridge regression is identical to the OLS objective function but with the addition of a regularisation term. Again, multiplying both components by half is to improve the simplicity where taking derivatives. The regularisation term is consists of a Euclidean norm term, $ \left\| \mathbf{w} \right \|^2$, and and a regularisation parameter $\lambda$.  The Euclidean norm terms computes the positive square root of the sum of squares of the weights, $ \left \| \mathbf{w} \right \|^2 := \sqrt{w_{1}^2+ w_{2}^2+...+w_{d}^2}$ and measures the complexity of a function. Functions with many large weights capable of producing highly non-linear functions will have large norms. The $\lambda$ parameter, a positive number that scales the norm and controls the amount of regularisation. When $\lambda$ is big, complex functions with large norms are heavily penalised, with $\lambda \left\|\mathbf{w}\right \|^2$  term significantly increasing the value of the objective function. Therefore, when $\lambda$ is big, minimising the objective function favours less complex functions and small weights.  As $\lambda$ approaches zero, large norms are less heavily penalised, with the product of $\lambda \left\|\mathbf{w}\right \|^2$ getting smaller and smaller,  allowing more complex functions with larger weights as optimal solutions to Equation 7. When $\lambda = 0$ the objective function is the same as the OLS model. 

The addition of regularisation forces the objective function to fit the data as closely as possible without creating functions that are too complex. Therefore, $\lambda$ can be used to control the bias-variance trade-off. As $\lambda$ get bigger the bias increases and variance decreases and vice versa. However, regularisation results in model weights that are biased towards zero, and thus (by design) the ridge estimate is a biased estimator. But as shown in the bias-variance trade-off, the increase in bias is out-weighted by lower variance and improved testing performance of the model. 

As was the case with linear regression, the aim is to find the vector of weights that minimise the ridge regression objective function:
\begin{equation}
    \min_{\mathbf{w }\in \mathbb{R}^d} \frac{1}{2} \left \|\boldsymbol{ y }- X\mathbf{w} \right \|^2 +\frac{\lambda}{2}\left\|\mathbf{w}\right \|^2
\end{equation}

Again this is calculated by taking the derivative of the objective function, setting it to zero, and solving for the weights:
\begin{equation}
    \frac{\partial S(\mathbf{w})}{\partial \mathbf{w}} = X^{T}(X\mathbf{w}-\mathbf{y}) + \lambda \mathbf{w} = 0 
\end{equation}
\begin{equation}
    \mathbf{w}^{*} = (X^{T}\!X + \lambda I_{n})^{-1}X^{T}\mathbf{y}
\end{equation}
For prediction we now have $\widehat{y_{*}} = \textbf{x}_{*}\mathbf{w}^{*}=\textbf{x}_{i}(X^{T}\!X + \lambda I_{n})^{-1}X^{T}\mathbf{y}$ where $I_{n}$ is the identity matrix (a square matrix in which all the elements of the principal diagonal are ones and all other elements are zeros) of dimensions $N \times N$. 

The above solution is termed the \textit{primal} solution of the ridge regression. The addition of $\lambda I_{n}$ ensures that when $\lambda >0$ the matrix $(X^{T}\!X + \lambda I_{n})$ is always invertible, and hence allows for stable computation. Secondly, the addition of lambda allows us to control the complexity of the optimal solutions. Optimising the primal involves solving the $d \times d$ system of equations given by $X^{T}X$,  with computational complexity $\mathcal{O}(d^{2}N)$ in the best case (the big O notation, $\mathcal{O}$, used to denote how the relative running time or space requirements grow as the input size grows). This complexity is extremely useful because even when the number of locations, $N$, is large, the dimensions of the explanatory variables dominate the primal solution. 

\section*{The Dual Solution and the Gram Matrix}
Interestingly, the primal form can be rewritten using the following matrix identity
\begin{equation}
   (X^{T}X + \lambda I_{n})X^{T} = X^{T}XX^{T}+\lambda X^{T} = X^{T}(XX^{T} +\lambda I_{n})
\end{equation}
Taking this rearrangement and plugging it back into the primal solution in Equation 5 gives a new equation for the weights and for model prediction:
\begin{equation}
    \mathbf{w}^{*} = X^{T}(XX^{T} +\lambda I_{n})^{-1}\mathbf{y}
\end{equation}
\begin{equation}
    \widehat{y_{*}} = \textbf{x}_{*}X^{T}(XX^{T} +\lambda I_{n})^{-1}\mathbf{y}
\end{equation}
This new equation can be further abstracted by defining  $\mathbf{\alpha} = (XX^{T} + \lambda I_{n})^{-1}\mathbf{y}$ such that the equations can be expressed as $\mathbf{w}^{*} = \sum_{i=1}^{N}\alpha_{i}\mathbf{{x}_{i}} = X^{T}\mathbf{\alpha}$. 

This derivation shows that the vector of weights, $\mathbf{w}$,  can be written as a linear combination of the training points where each $\textbf{x}_{i}$ represents a row of the design matrix corresponding to the explanatory variables at a location. For any new observation the output is given by $ \widehat{y_{*}} = \mathbf{{x}_{*}}\sum_{i=1}^{N}\alpha_{i}\mathbf{{x}_{i}} = \mathbf{x}_{*}X^{T}\mathbf{\alpha}$. This solution is termed the \textit{Dual} solution of the ridge regression and the variables $\alpha \in \displaystyle \mathbb{R}^N$ are called dual variables.  Rather than finding the optimal weights for each of the $d$ explanatory variable (the column of the design matrix), we find the appropriate dual variable for each of the $N$ locations (the rows of the design matrix). In contrast to the primal, the dual solution requires inverting an $N \times N$ dimensional matrix, $XX^{T}$, with complexity $\mathcal{O}(N^{3})$ in the worst case.  
 
Given, that the dual has been derived directly from the primal it is easy to see that the solutions are equivalent, with the two solutions said to exhibit \textit{strong duality} \cite{boyd2004convex}. However, the dual form can be derived directly from ridge regression objective function independent of the primal by expressing the problem as a constrained minimisation problem and solving using the Lagrangian (Supplementary Equations 1).  
 
Given the much higher complexity of the dual solution, it is not immediately obvious why this solution would ever be useful. However, a property of dual solution in Equation 12 is that it requires computing $XX^{T}$, with the resulting matrix a symmetric positive semidefinite matrix called the Gram matrix, $G$. $G$ contains all of the pairwise inner product of inputs across all $N$ locations. An inner product is a way to multiply vectors together, with the result of this multiplication being a scalar measure of the vectors \emph{similarity}. Explicitly, the inner product of two vectors $\textbf{x}$ and $\textbf{z} \in \mathbb{R}^{d}$ is given by:
 \begin{equation}
\begin{split}
       \left \langle \mathbf{x} , \mathbf{z} \right \rangle & =  \left \langle(x_{1}, x_{2}, x_{3},..., x_{d}),(z_{1}, z_{2}, z_{3},..., z_{d})  \right \rangle  \\
       & = x_{1}z_{1} + x_{2}z_{2}+ x_{3}z_{3} + ... + x_{d}z_{d} 
\end{split}
\end{equation}
where $\left \langle \cdot ,\cdot \right \rangle$ is used to signify the inner product.  Therefore, the Gram matrix can be thought of as containing the similarity between all pairs of inputs. Indeed, if these vectors are centred random variables, $G$ is approximately proportional to the covariance matrix. This notion of similarity is central in spatial analysis where we want to leverage the fact the points close to each other in space are similar.

As an example, the entry at row $i$ column $j$ of matrix $G$ represents the inner product between the vectors of explanatory variables at location $i$ and $j$ respectively (corresponding to rows $i$ and $j$ in the design matrix).  This is written as $g_{ij} = \left \langle \mathbf{x}_{i}, \mathbf{x}_{j} \right \rangle$.  The full gram matrix is given by $G = \left \langle X, X \right \rangle$. Therefore, in the dual solution, we can substitute $XX^{T}$ for the $G$, and $\textbf{x}_{*}X^{T}$ with $\mathbf{g}(\mathbf{x}_{*}, X)$,  the inner product between a new data point and the training points, ($\left \langle \mathbf{x}_{*}, X \right \rangle$) giving:
\begin{equation}
\begin{split}
    & \alpha = (G +\lambda I_{n})^{-1}\mathbf{y} \\
    & \widehat{y_{*}} = \mathbf{g}(\mathbf{x}_{*}, X) \alpha
\end{split}
\end{equation}
However, the question remains, how can we use this useful construction of the Gram matrix to model non-linear functions?
 
\section*{Non-Linear Regression}
Up to this point, the equations have only represented linear models, where the outputs are assumed to be some linear combination of the inputs. However, many problems cannot be adequately described in purely linear terms. One approach to introduce non-linearity to the model is to transform the explanatory variables using non-linear transformations, such that the output is described as a linear combination of non-linear terms. For example, rather than a weighted sum of linear terms (i.e. $x_{1}+x_{2}+x_{3}$), we  may instead use terms with exponents, logarithms or trigonometric functions (i.e. $exp(x_{1})+log(x_{2})+sin(x_{3})$). Transforming the inputs rather changing the model allows us to maintain all of the convenient maths we have derived for linear models to create non-linear ones. 

Mathematically, the input data is said to be projected from a linear input space to some new, and potentially non-linear, space. The projecting of data is termed a (non-linear) \textit{feature mapping} and the new space to which the data is mapped is called the \textit{feature space}. Figure 2 is an example of a mapping to a feature space such that the output can be expressed in linear terms (Supplementary Code 2).
\begin{figure}[ht]
  \includegraphics[width=\textwidth,height=\textheight,keepaspectratio]{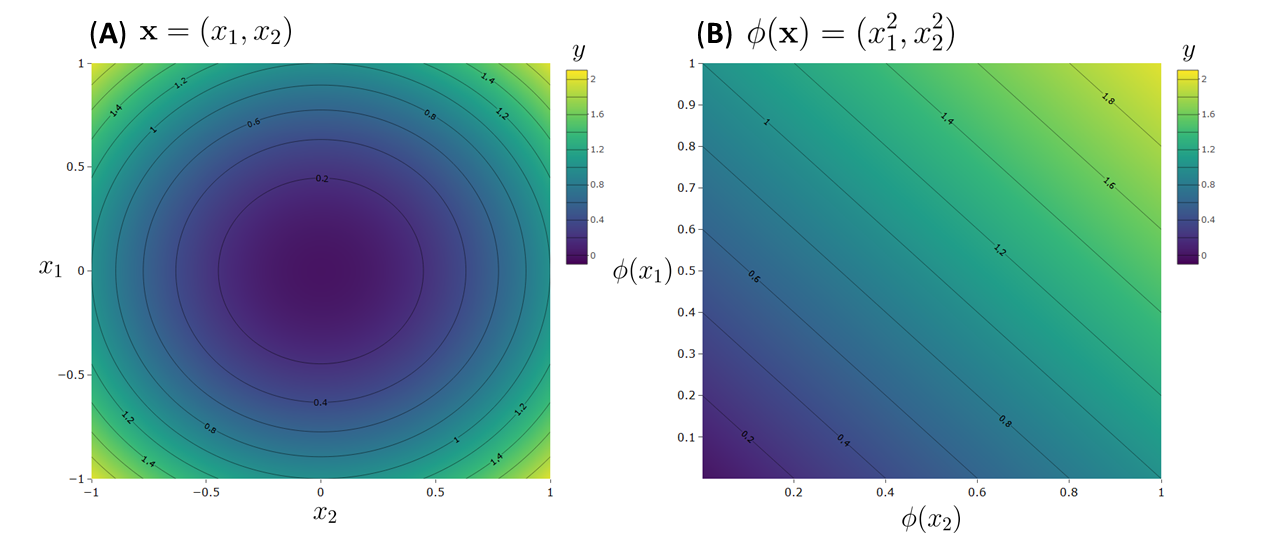}
  \caption{An example of non-linear feature mapping, where the data is mapped for (\textbf{A}) an input space in which the problem is non-linear into a new feature space (\textbf{B}) in which the outputs can be described as a linear combination of the inputs. }
  \label{fig.2}
\end{figure}

Generally, a mapping is denoted by $\phi$ (or $\Phi$ when applied to a matrix). The general form of a feature mapping is given by:
\begin{equation}
    \phi : \mathbf{x}_{i} \in \mathbb{R}^d  \mapsto  \phi (\mathbf{x}_{i}) \in \mathbb{R}^D
\end{equation}
\begin{equation}
    \Phi : X \in \mathbb{R}^{N \times d}  \mapsto  \Phi(X) \in \mathbb{R}^{N \times D}
\end{equation}
The vector, $\textbf{x}_{i}$, is mapped from a vector of length $d$ to a vector of length $D$ denoted as  $\phi(\textbf{x}_{i})$. Applying this mapping to the entire design matrix gives a new design matrix of explanatory variables in the feature space, $\Phi(X) \in \mathbb{R}^{N \times D}$.  A mapping can project to into higher-dimensional ($d < D$) or lower-dimensional ($d > D$) space, although in the context of regression, \textit{lifting} to a higher dimension is more common. As an explicit example, consider the following mapping:
\begin{equation}
    \phi : \mathbf{x} =(x_{1},x_{2})  \mapsto  \phi (\mathbf{x}) = (x_{1}^2, \sqrt2 x_{1}x_{2}, x_{2}^2)
\end{equation}
This mapping \textit{lifts} the data from 2-dimensions into 3-dimensions ($\mathbb{R}^{2} \mapsto \mathbb{R}^{3}$). The same set of equations can be used to solve a linear model in feature space after the mapping. For example, the primal for ridge regression is now given by:
\begin{equation}
\min_{\mathbf{w}\in \mathbb{R}^D} \left \|\boldsymbol{y}- \Phi(X)\mathbf{w} \right \|^2 +\frac{\lambda}{2}\left\|\mathbf{w}\right \|^2
\end{equation}
\begin{equation}
\mathbf{w}^{*} = (\Phi(X)^{T}\!\Phi(X)+ \lambda I_{n})^{-1}\Phi(X)^{T}\mathbf{y}
\end{equation}
\begin{equation}
  \widehat{y_{*}} = \phi(\textbf{x}_{*})\mathbf{w}^{*}=\phi(\textbf{x}_{*})(\Phi(X)^{T}\!\Phi(X) + \lambda I_{n})^{-1}\Phi(X)^{T}\mathbf{y}  
\end{equation}
All that was required to derive these equations is to substitute the design matrix in the original input space, $X$, with the new design matrix in the feature space, $\Phi(X)$ (with the equivalent mapping for new input points, $\phi(\textbf{x}_{*})$). This substitution also applies to standard linear regression and the ridge regressor dual but are omitted for brevity. The primal solution now requires solving for the weights in $R^{D}$. Therefore, it is easy to see how a very high dimensional mapping ($D \gg d$ and $D > N$) solving the primal will computationally very difficult. Thankfully, due to our dual ridge solution, this overdetermined situation does not present a computational problem - no matter how many terms we add, the complexity will still be $\mathcal{O}(N^3)$.  

However, it is rarely apparent \textit{a priori} what is the most appropriate mapping to apply to the data. Therefore, while the dual allows for very high dimensional feature mapping, the question remains, which mapping should we use? How many terms should be added? How do we capture interactions between terms? A brute force approach can quickly become combinatorially large. Given that we can limit model complexity using a ridge regularisation, the ideal situation would be to be able to define a large, if not infinite mapping capable of nearly any function. We can do this using kernels.

\section*{Kernel Methods}
As highlighted earlier, an ideal feature mapping would be to a large or infinite feature space to which we can then apply regularisation. The dual solution ensures that we only need to solve for the same number of dual variables regardless of the dimensionality of the feature mapping (remember dual variables are in $R^{N}$ with complexity is $\mathcal{O}(N^3)$). However, this raises an interesting question: can we solve ridge regression in an infinite-dimensional feature space?
\begin{equation}
\begin{split}
    & \phi : \mathbf{x}_{i} \in \mathbb{R}^d  \mapsto  \phi (\mathbf{x}_{i}) \in \mathbb{R}^\infty \\ 
    & \Phi : X \in \mathbb{R}^{N \times d}  \mapsto  \Phi(X) \in \mathbb{R}^{N \times \infty}
\end{split}
\end{equation}
We do not want to (and can't) compute all the infinite terms required for explicit infinite-dimensional feature mapping. To work with these infinite dimensional spaces, we need to move away from explicit to implicit feature maps. To arrive at an implicit map, consider solving the dual using the aforementioned feature mapping in Equation 18. The dual solution requires computing the inner product the inner product between all pairs of inputs in feature space, such that the inner product between any two input vectors, $\textbf{X}$ and $\textbf{Z}$, first requires their mapping to feature space:

\begin{equation}
\begin{split}
    & \phi :\mathbf{x} =(x_{1},x_{2})  \mapsto  \phi (\mathbf{x}) = (x_{1}^2, \sqrt2 x_{1}x_{2}, x_{2}^2) \\
    & \phi  :\mathbf{z} =(z_{1},z_{2})  \mapsto  \phi (\mathbf{z}) = (z_{1}^2, \sqrt2 z_{1}z_{2}, z_{2}^2)
\end{split}
\end{equation}

Then computing the inner product between the vectors in feature space:

\begin{equation} 
\begin{split}
\left \langle \phi (\mathbf{x}),\phi (\mathbf{z}) \right \rangle & = \left \langle (x_{1}^2, \sqrt2 x_{1}x_{2}, x_{2}^2) ,(z_{1}^2, \sqrt2 z_{1}z_{2}, z_{2}^2)  \right \rangle \\
&= x_{1}^2z_{1}^2 + 2 x_{1}x_{2}z_{1}z_{2} +  x_{2}^2z_{2}^2
\end{split}
\end{equation} 

This calculation of the inner product gives a scalar value of the similarity between the two vectors in feature space but required explicitly computing each of the feature mappings of the two vectors before taking their inner product. What is interesting is that the equation for the inner-product in feature space (Equation 24) takes the form of a quadratic equation between $\textbf{x}$ and $\textbf{z}$. Factorising the quadratic gives:
\begin{equation} 
\begin{split}
  \left \langle \phi (\mathbf{x}),\phi (\mathbf{z}) \right \rangle & = x_{1}^2z_{1}^2 + 2 x_{1}x_{2}z_{1}z_{2} +  x_{2}^2z_{2}^2 \\
  & = (x_{1}z_{1} + x_{2}z_{2})^{2} \\
  & = \left \langle \mathbf{x},\mathbf{z} \right \rangle^{2}
 \end{split}
\end{equation}
  
We have just arrived at a new way of computing the inner product in feature space. The inner product between the vectors in feature space is equal to the square of the inner product in input space. The beauty of this result is our new function for the inner product in feature space does not require computing the explicit mapping of our data (does not require computing $\Phi(X)$). We have already established that solving the dual only requires the inner product. Therefore, a solution to the dual can be obtained without ever computing and storing the explicit feature mapping as long as we have a function to directly compute the inner product in feature space. 

The ability for linear models to learn a nonlinear function while avoiding an explicit mapping is a famed result in machine learning known as the kernel trick \cite{Boser1992} with the functions that compute the inner product in feature space are called \textit{kernels functions}. Kernels, therefore, perform the mapping $k:\mathcal{X} \times \mathcal{X} \mapsto \mathbb{R}$ between inputs $x,z \in \mathcal{X}$. The resulting matrix of inner products generated by the kernel function is termed the \textit{kernel matrix} (and can be thought of as the Gram matrix in feature space). We can represent the kernel matrix by $K$ (or $k$ for vectors) such that:
\begin{equation}
\begin{split}
    K(X, X) & = \left \langle \Phi(X),\Phi(X)\right \rangle = \Phi(X)\Phi(X)^{T} \\
    k(\mathbf{x}_{*}, X) & = \left \langle \phi(\mathbf{x}_{*}),\Phi(X)\right \rangle = \phi(\mathbf{x}_{*})\Phi(X)^{T}
\end{split}
\end{equation}
The dual equations can now be written in terms of kernels given by:
\begin{equation}
     \widehat{y_{*}} = k(\textbf{x}_{*},X)(K(X,X) + \lambda I_{n})^{-1}\mathbf{y}
\end{equation} 
 
This equation encapsulates all the theory we have developed thus far. Using the squared loss, we can compute in closed form the optimal solution for the weights. Using the kernel trick, we can project our input data implicitly into an infinite-dimensional feature space. Using regularisation through the ridge penalty, we can prevent our solution from overfitting. Thus, we have derived a powerful non-linear model using no more same equations than we would have used for a linear model.

For the specific example of the mapping in Equation 23, the kernel function is given by $K(\mathbf{x},\mathbf{z}) = \left \langle \mathbf{x},\mathbf{z} \right \rangle^{2}$ and represent a second-degree polynomial kernel. A more general formula for a degree-$p$ polynomial kernel is given by $K_{\theta, p}(\mathbf{x},\mathbf{z})= \left \langle \mathbf{x},\mathbf{z} + \theta \right \rangle^{p}$ (for values of $p \in \mathbb{Z}^{+}$ and $\theta \in \mathbb{R}_{\geq 0}$), where $\theta$ is a free parameter that allows control over the influence of higher-order and lower-order terms in the polynomial in the kernel function. However, there are many kernel functions; an example of an infinite dimensional kernel is the popular and widely used squared exponential kernel:
\begin{equation}
    K_{\ell, \sigma}(\mathbf{x},\mathbf{z})=\sigma^2\exp\left(-\frac{(x - z)^2}{2\ell^2}\right) 
\end{equation}
Where $\ell$ called the length scale, controls the distance over which the range over which the kernel similarly operates and $\sigma$ determines the average distance from the mean. The proof that a squared exponential kernel gives rise to an infinite feature mapping uses a Taylor series expansion of the exponential, shown in Supplementary Equations 2. 

\section*{The Big $N$ Problem} 
It is evident that the combination of the dual estimator and the kernel function is a powerful tool capable of extending linear models to handle non-linear problems. One of the primary motivations for considering the dual was that it required computing and inverting an $N \times N$ dimensional matrix rather than the $d \times d$ dimensional matrix of features in the primal. Even as $d$ grows infinitely large, the kernel matrix remains of constant size and involves solving for the same number of dual variables. Historically this led to the widespread adoption of kernel methods (kernel ridge regression, support vector machines, Gaussian processes etc.) to solve difficult problems on small datasets. However, the dual still requires inverting and storing the kernel matrix, which for $N$ observations will be $\mathcal{O}(N^{3})$ complexity and need $\mathcal{O}(N^{2})$ storage. Given only a few thousand points, these polynomial costs rapidly outstrip computational power. 

A plethora of methods have been developed to aid the scaling of kernels methods to large datasets. Broadly, these methods aim to find smaller/simpler matrices that are good approximations of the full kernel matrix. The three major techniques are low-rank approximations, sparse approximations and spectral methods. Low-rank approximations of a matrix aim to find smaller representations of the kernel matrix that contains all (or nearly all) of the information in the full kernel \cite{Bach2005}. For example, the popular Nystr{\"o}m approximates the full kernel matrix through a subset of its columns and rows \cite{williams}. In comparisons, sparse methods aim to find representations of the matrix that are mostly zeros because there exist efficient algorithms for the storage of and computation with such matrices \cite{rue2005gaussian, straeter1971extension, saad1986gmres}. One of the best examples is the sparse matrix generated when modelling spatial data as a Gaussian Markov random field (GMRF) that are solutions to Stochastic Partial Differential Equation (SPDE) \cite{Lindgren, whittle1954stationary, whittle1963stochastic}. However, the remainder of this paper will focus on a new, exciting subset of spectral methods called random Fourier features (RFF).

\section*{Random Fourier Features}
RFF and other spectral method utilise the characterisation of the kernel function through its Fourier transform. A Fourier transform allows for the decomposition of any function into the periodic functions of different frequencies that make it up. The central idea behind these methods is that a good approximation of the kernel in the frequency domain (where the function described in terms of the periodic functions that make it up) will naturally yield a good approximation to the kernel function in its original domain.

All spectral methods are based on the same mathematical foundation; specifically, the celebrated \textit{Bochner's theorem} \cite{bochner1949fourier}. Loosely, Bochner's theorem states that a shift-invariant kernel functions (where the output of the kennel is only dependent on the difference between the inputs and not the explicit values of the input themselves) $k(x_{1},x_{2}) = k(\delta)$ for $\delta=|x_{1}-x_{2}|$, can be expressed through a Fourier transform \cite{Rudin1990}:  
\begin{equation}
    k(x_{1}-x_{2}) = \int_{R^{d}}e^{i\omega^{T}(x_{1}-x_{2})} \mathbb{P}(\omega ) \; d\omega
\end{equation}

If we apply Euler's identity to the exponential and ignore the imaginary component of Equation 29 we can express this integral as:
\begin{equation}
    k(x_{1}-x_{2}) = \int_{R^{D}} 
\begin{pmatrix}
 \cos(\omega^T x_1)\\
 \sin(\omega^T x_1)
\end{pmatrix}^T
\begin{pmatrix}
 \cos(\omega^T x_2)\\
 \sin(\omega^T x_2)
\end{pmatrix}
\mathbb{P}(\omega ) \; d\omega
\end{equation}

This real part of Bochner's theorem computed by projecting the data onto the line drawn by $\omega$ (given by $\omega^T x_1$ and $\omega^T x_2$), pass these projections through $\cos$ and $\sin$ and stack them together. To understand why this process works consider the two key components, $\omega$ and its distribution $\mathbb{P}(\omega )$. The variable $\omega$ is the frequency of the periodic functions. The distribution of these frequencies, $\mathbb{P}(\omega)$, is called the spectral density of the kernel and gives the 'importance' of a given frequency, $\omega$, in constructing the kernel function. This is visualised in Figure 3, showing different spectral densities (Figures 3A,C,E,G,I) and the resulting functions produced by sampling from the kernel generated by each spectral density (Figures 3B,D,F,H,J). The code for sampling the spectral densities and generating functions is given in Supplementary Code 3. 

\begin{figure}[ht]
  \includegraphics[width=\textwidth,height=\textheight,keepaspectratio]{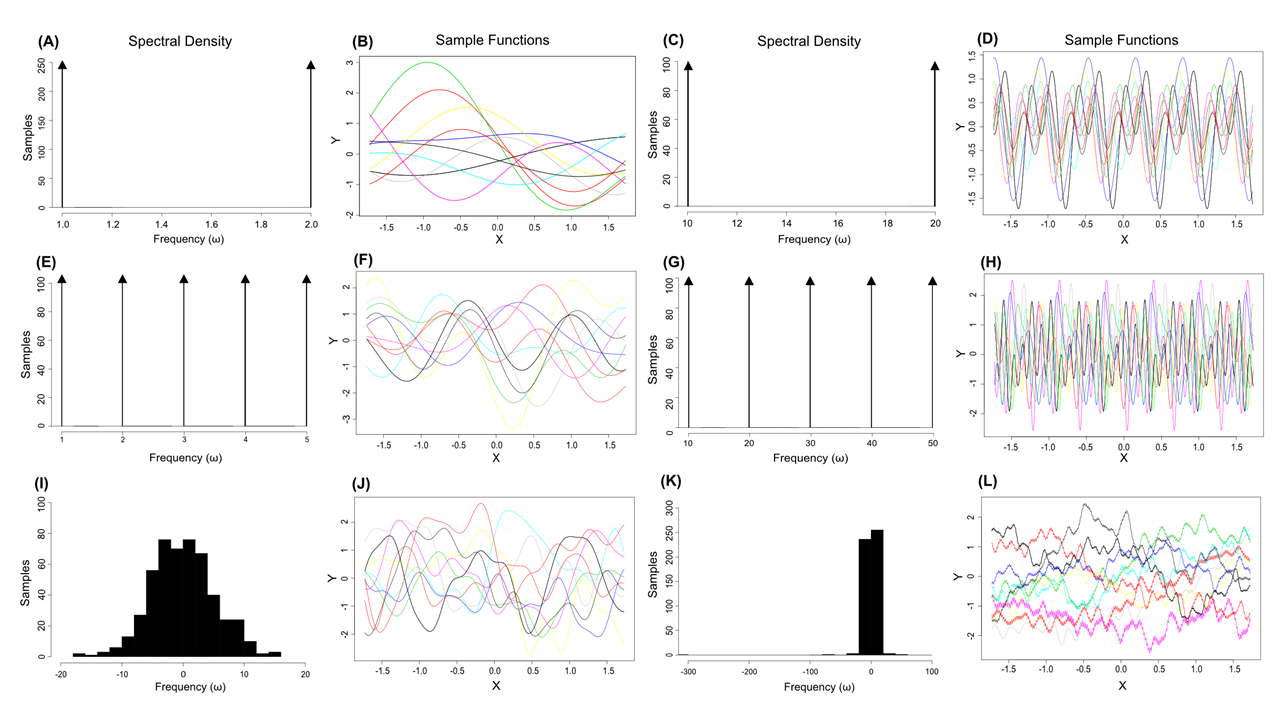}
  \caption{Power spectral densities and the functions produced by sampling from the resulting kernel. The spectral densities in A,C,E,G correspond to sampling from delta functions (arrowheads) such that the sampled frequencies, $\omega$, can only take the point values corresponding to each delta function.}
  \label{fig.3}
\end{figure}

In figure 3A the spectral density is composed of two delta functions, such that samples frequencies ($\omega$) can only take values of $1$ or $2$. The functions generated by sampling from this spectral density show strong periodicity and closely resemble the standard trigonometric functions with corresponding frequencies (Figure 3D). When the frequencies of the two delta functions are increased to $\in\{10,20\}$, the functions are again highly cyclical but, due to their higher frequencies, have rougher sample paths and a much smaller period (small changes in $x$ can cause greater changes in $y$) (Figure 3C,D). By expanding the spectral density to 5 peaks, the sample paths show considerably more variation due to the inclusion of a larger variety of frequencies (Figure 3E,F and 3G,H). Finally figures 3H and 3J show samples functions generated by sampling frequencies form a Gaussian and Cauchy distribution respectively. The Gaussian spectral density corresponds to a spectral density of the squared exponential kernel (Gaussian kernel) and gives rise to smooth sample functions with a tremendous amount of variety when compared to the simpler spectral densities (Figure 3J). The Cauchy distribution corresponds to a spectral density generated by the Fourier transform of the Laplacian kernel and generates functions with a high degree of roughness (Figure 3L) due to the inclusion of very high frequencies in the long tails of the distribution (Figure 3K). Table 1 below shows the resulting power spectral densities generated by some common shift-invariant kernels. It should also be noted that empirical spectral densities can be used \cite{TON201859, Lazaro-Gredilla2010, Quinonero-candela2005}, and non-stationary extensions of Bochner's theorem exist \cite{TON201859, yaglom2012correlation} (discussed in Non-stationary and arbitrary Kernel Functions section).

However, a major problem is that evaluating the integral in Equation 30 requires integrating over the infinite set of all possible frequencies. To get around this, we can approximate this infinite integral by a finite one using Monte Carlo integration. In Monte Carlo integration the full integral of a function is approximated by computing the average of that function evaluated at a random set of points. Therefore, for RFF, rather than an infinite set of frequencies, the spectral density is approximated by averaging the sum of the function evaluated at random samples of $\omega$ drawn from the spectral density. The more samples that are evaluated, the closer the approximation gets to the full integral. Indeed, one of the best properties of random Fourier features is that of uniform convergence, with the Monte Carlo approximation of the entire kernel function converging to the full kernel uniformly (rather than pointwise).

\begin{table}[ht]
\centering
\begin{tabular}{|p{2cm}|l|l|}
\hline
\textbf{Kernel} & \multicolumn{1}{c|}{\textbf{Kernel Function}, $K(x_{1}-x_{2})$} & \multicolumn{1}{c|}{\textbf{Power Spectral Density} , $\mathbb{P}(\omega)$} \\ \hline
\begin{tabular}[c]{@{}l@{}}\textit{Squared}\\ \textit{Exponential}\end{tabular} & $\sigma^2\exp\left(-\frac{(x_{1}-x_{2})^2}{2\ell^2}\right)$ & $(2\pi)-\frac{D}{2}\sigma^{D}\exp\left(-\frac{\sigma^{D}\left \| \omega \right \|_{2}^{2}}{2}\right)$ \\ \hline
\textit{ Mat{\'e}n} & \small{$\frac{2^{1-\lambda}}{\Gamma(\lambda)}\left (\frac{\sqrt{2\lambda \left \| x_{1}-x_{2} \right \|_{2}}}{\sigma}  \right )^{\lambda} K_{\lambda} \left (\frac{\sqrt{2\lambda \left \| x_{1}-x_{2} \right \|_{2}}}{\sigma}   \right )$} & \small{$\frac{2^{D+\lambda}\pi^{\frac{D}{2}} \Gamma(\lambda+\frac{D}{2})\lambda^{\lambda}}{\Gamma(\lambda)\sigma^{2\lambda}} \left (\frac{2\lambda}{\sigma^{2}}+4\pi^{2}\left \| \omega \right \|^{2}_{2} \right )^{-(\lambda+\frac{D}{2})}$}  
\\ \hline
\textit{Cauchy} & $\prod_{i=1}^{D}\frac{2}{1 +x_{i}^{2}} $ & $ \exp (-\left \| \omega  \right \|_{1})$ \\ \hline
\textit{Laplacian} &  $\exp \left (-\sigma\left \| x_{1}-x_{2} \right \|_{1}  \right )$ &  $\left (\frac{2}{\pi} \right )^{\frac{D}{2}}\prod_{i=1}^{D}\frac{\sigma}{\sigma^{2} +\omega_{i}^{2}}$  \\ \hline
\end{tabular}
\caption{Common shift-invariant kernels and their associated spectral densities}
\label{table:1}
\end{table}

Therefore, the infinite integral in Equation 30 can be converted to a finite approximation by taking multiple independent samples from the power spectral density, $\mathbb{P}(\omega)$, and computing the Monte Carlo approximation to the kernel $k(x_{1}-x_{2})$ via:
\begin{equation}
    k(x_{1}-x_{2}) = \frac{1}{m}\sum_{j=1}^{m} 
\begin{pmatrix}
 \cos(\omega_j^T x_1)\\
 \sin(\omega_j^T x_1)
\end{pmatrix}^T
\begin{pmatrix}
 \cos(\omega_j^T x_2)\\
 \sin(\omega_j^T x_2)
\end{pmatrix}
= \Phi_{\tiny{\textup{RFF}}}(x_{1})\Phi_{\tiny{\textup{RFF}}}(x_{2})^T 
\;\;\;\;\;\;\;,\left \{\omega\right\}_{j=1}^{m} \overset{i.i.d.}{\sim } \mathbb{P}(\omega)
\end{equation}

Given that the spectral densities are probability distributions, it is often reasonably trivial to sample frequencies from them. For example, generating the frequencies for approximating a squared exponential is as simple as independently sampling $\omega$'s from a Gaussian distribution (Table 1). Equation 31 is truly an astoundingly condensed result and can be written in its entirety in just 4 lines of R-code:\newline

\begin{lstlisting}[language=R, caption=Example of creating random Fourier features to approximate a Gaussian kernel matrix.]
# data matrix X of dimensions (N x d), number of features m
Omega = matrix(rnorm(m*ncol(X)), m) # squared exponential kernel
Proj = x %*% t(Omega) # projection
Phi = cbind(cos(Proj), sin(Proj)) / sqrt(m) # basis
K = Phi %*% t(Phi) # kernel matrix
\end{lstlisting}

The RFF approach results in an approximation of the whole kernel matrix. Therefore, it is different from other low-rank methods that try to approximate parts of the full kernel matrix. Using the Woodbury matrix inversion formula we can reduce the complexity of inverting the whole kernel matrix from $\mathcal{O}(N^{3})$ to $\mathcal{O}(m^{3})$ (where $m$ is the number of $i.i.d.$ samples from $\mathbb{P}(\omega)$) to solve the dual \cite{MR0038136}. However, one of the key observations about RFF's is that the random basis matrix $\Phi_{\tiny{\textup{RFF}}}$ defines a function space of its own! Technically a function space that is dense in a reproducing Hilbert space - the same space of functions from our kernel matrix. We can define this feature space as:
\begin{equation}
    \Phi_{\tiny{\textup{RFF}}}(x) = 
\begin{pmatrix}
 \cos(\omega_j^T x)\\
 \sin(\omega_j^T x)
\end{pmatrix} \;\; \; \; \; \; \left \{\omega\right\}_{j=1}^{m} \overset{i.i.d.}{\sim } \mathbb{P}(\omega) 
\end{equation}
This can be written using matrix notation as $\Phi(X)  = [cos(X\Omega^{T}) \;sin(X\Omega^{T})] \in \mathbb{R}^{N \times 2m}$, where matrix $X\in \mathbb{R}^{N \times d}$ is the design matrix and $\Omega \in \mathbb{R}^{m \times d}$ is the frequency matrix with rows corresponding a sampled $\omega$ ($[\omega_{1},...,\omega_{m}]$). See Supplementary Equations 3 for a more comprehensive walk-through of the steps from Bochner's theorem to the RFF equation.  

We can use the $\Phi_{\tiny{\textup{RFF}}}(X)^{T}\Phi_{\tiny{\textup{RFF}}}(X) \in \displaystyle \mathbb{R}^{m \times m}$ matrix to solve the primal. The resulting linear model takes the form $y \sim \Phi_{\tiny{\textup{RFF}}}(X)\textbf{w}$. Therefore, for a spatial data set, the method only requires mapping the explanatory variables using RFF and fitting a linear model. Therefore, the ubiquitous linear model is converted to a much more expressive form, while retaining all the desirable mathematical properties that make linear models so popular.

The theoretical properties of RFF estimators are still far from fully understood. The seminal paper by Rahimi and Recht \cite{Rahimi2007} showed that every entry in our kernel matrix is approximated to an error of $\pm \epsilon $ with $m = \frac{\log(N)}{\epsilon^{2}}$ Fourier features. A more recent result shows that only $\sqrt{N}\log(N)$ features can achieve the same learning bounds as full kernel ridge regression with squared loss \cite{rudi2017generalization}. This is particularly relevant for large datasets. For example, given $100,000$ data points, we would only need $~3600$ features to achieve the generalisation error as if we had used all points.

\section*{Beyond the General Linear Model} 
Throughout this paper, we have focused on Gaussian likelihoods/Squared loss, but Fourier bases can be used with any loss function. For example, when performing classification with binary data, i.e. $y\in\{0,1\}$ the cross-entropy loss (also known as log loss) can be used, given by:
\begin{equation}
S(\mathbf{w}) = -\left((y\log(\varphi^{-1}(X\mathbf{w}))+(1-y)\log(1-\varphi^{-1}(X\mathbf{w})) \right)
\end{equation}

where $\varphi$ is the Sigmoid or Logit function. Another example is the use of a Poisson likelihood to model count data \cite{cameron2013regression}. More broadly, generalised linear models (GLM) encompass the extension of linear regression to response variables that have a restricted range of potently values (such as binary or count data) or non-Gaussian error \cite{nelder1972generalized}. This generalisation is achieved by letting the linear be related to the response variable via a link function. The link function ensures that the model's estimated responses are on the correct range and have the appropriate error structure but is still a function of the weighted linear sum of explanatory variables. Thus, we can still apply feature mapping including RFF and can continue to fit these models using maximum likelihood.  

These models can easily be extended to include uncertainty through Bayesian inference. For example, Bayesian linear regression is achieved by assuming that both the response variables and the model parameters are drawn from distributions, with the aim to find the posterior distribution of the model parameters given the input and output variables. By specifying a Gaussian likelihood, with a Gaussian prior on our coefficients, in expectation, our posterior is the same as that of ridge regression. We can evaluate full posterior uncertainty by sampling from the posterior distribution.

\section*{Toy Example of Random Fourier Features for Spatial Analysis}
As an example, we simulate a non-linear spatial regression problem. The code for this example is provided in Supplementary Code 4. A set of random points in space is generated, such that each location has unique coordinates (longitude and latitude). Each location has a response variable generated from the same latent function as in Figure 2 with added Gaussian noise ($\textbf{y} = \textbf{x}_{1}^{2}+\textbf{x}_{2}^{2}+\epsilon$ where $\epsilon \sim N(\mu=0,\sigma^{2}=1)$). In figure 4a we show random 500 points drawn from the spatial process (the code can easily be changed to any function of the user’s choice). Note that in the provided code generates points at random and therefore a user's results may differ from the exact results shown here. 

The simulated data were used to train three models, a linear regression model, a non-linear, kernel regression model and a kernel ridge regression model (KRR). Both the kernel regression and kernel ridge regression models use RFFs to approximate a Gaussian kernel approximation (user can specify the number of features). We assume, as is common for nearly all real-world spatial processes, that we only observe a subset of all possible locations. Therefore, models are trained on only 20\% of all the generated points, shown in Figure 4b. The remaining data not used to training is used for testing.

\begin{figure}[ht]
\centering
  \includegraphics[width=\textwidth,height=\textheight,keepaspectratio]{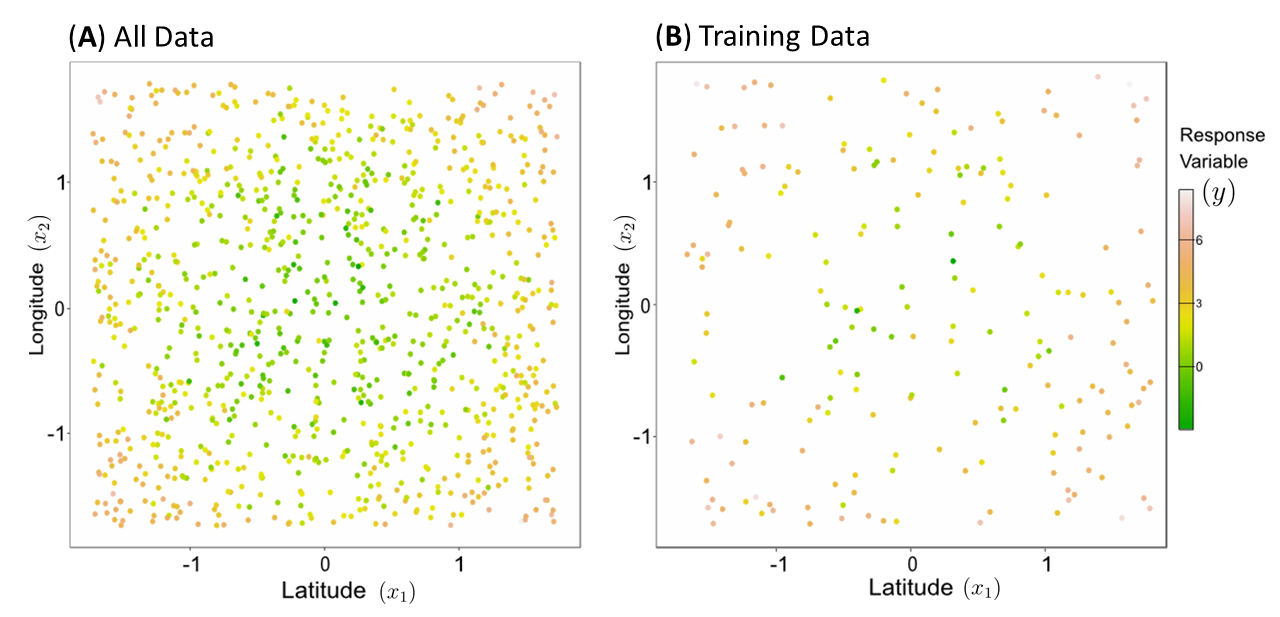}
  \caption{\textbf{(A)} All 500 points generated from the latent spatial process given by $\mathbf{y} = \mathbf{x}_{1}^2 + \mathbf{x}_{2}^2+\epsilon$  where $\epsilon \sim N(\mu=0,\sigma^{2}=1)$), and (\textbf{B}) the subset of data points used to train the regression models.}
  \label{fig.4}
\end{figure}

Each model was trained using the same training data and their predictive performance measured by MSE for the testing data. For both kernel methods (kernel regression and kernel ridge regression), 100 Fourier features are used. For KRR, k-fold cross-validation was used to find the optimal regularisation parameter ($\lambda_{ridge}$). The training and testing performance of the three models are shown in Table 2. As expected, the non-linear nature of the latent function results in very poor performance of the linear model with large training and testing error. The difference between the kernel regression (without regularisation) and KRR (with regularisation) is indicative of bias-variance trade-off discussed earlier. 

The Kernel regression model has excellent training performance, with the infinite feature space of the Gaussian kernel permitting the fitting of highly complex functions. However, in the absence of regularisation, the kernel regression model greatly overfits the training data resulting in poor testing performance. If we consider the bias-variance trade-off, the kernel regression model has a very low bias, but high variance. In comparison,  the regularisation in kernel ridge regression help to prevent this overfitting by penalising the overly complex models. After training,  the KRR model (with $\lambda_{ridge}=3.98$) has marginally higher training error than the kernel regression model but less than half the testing error. The regularisation has increased the KRR model's bias, but this is outweighed by the reduction in variance, corresponding to a small decrease the training accuracy but significant improvement in testing performance. Note, in this example the latent function is fairly simple (a 2D quadratic); therefore the number of features required for good performance is low. 

\begin{table}[ht]
\centering
\begin{tabular}{|l|c|c|}
\hline
\textbf{Model}                                                                                & \textbf{Training Error (MSE)}& \textbf{Testing Error  (MSE)}\\ \hline
\textit{Linear}                                                                               & 2.26                 & 2.73                 \\ \hline
\begin{tabular}[c]{@{}l@{}}\textit{Kernel Regression}\\ \textit{(Gaussian Kernel)}\end{tabular}    & 0.64                 & 2.70                 \\ \hline
\begin{tabular}[c]{@{}l@{}}\textit{Kernel \textbf{Ridge} Regression} \\ \textit{(Gaussian Kernel)}\end{tabular} & 0.88                 & 1.19                 \\ \hline
\end{tabular}
\caption{Training and testing performance of different models}
\label{table:2}
\end{table}
 
The importance of the bias-variance trade-off further illustrated by comparing how the training and testing performance of kernel regression and KRR vary with the number of Fourier features increases with, shown in Figure 5.  Increasing the number of sampled Fourier bases increases the model's ability to fit the training data and results in a steady reduction in training error kernel regression (Figure 5a, blue line). In comparison, KRR has a higher than the training error than kernel regression that remains constant even with additional features (Figure 5a, red line). As the number of features increases the kernel regression model increasingly overfits the training data resulting in poor testing performance (Figure 5b, blue line).Importantly, the testing performance of KRR is significantly lower than kernel regression and remains low even with additional features (Figure 5b, red line). The regularisation in KRR constrains model complexity, such that as more features are added the regularisation prevents overfitting by increasing the magnitude of the regularisation parameter, $\lambda_{ridge}$ (Figure 5b, inset). As a result, KRR maintains a good bias-variance trade-off across all numbers of features. 

\begin{figure}[h]
  \includegraphics[width=\textwidth,height=\textheight,keepaspectratio]{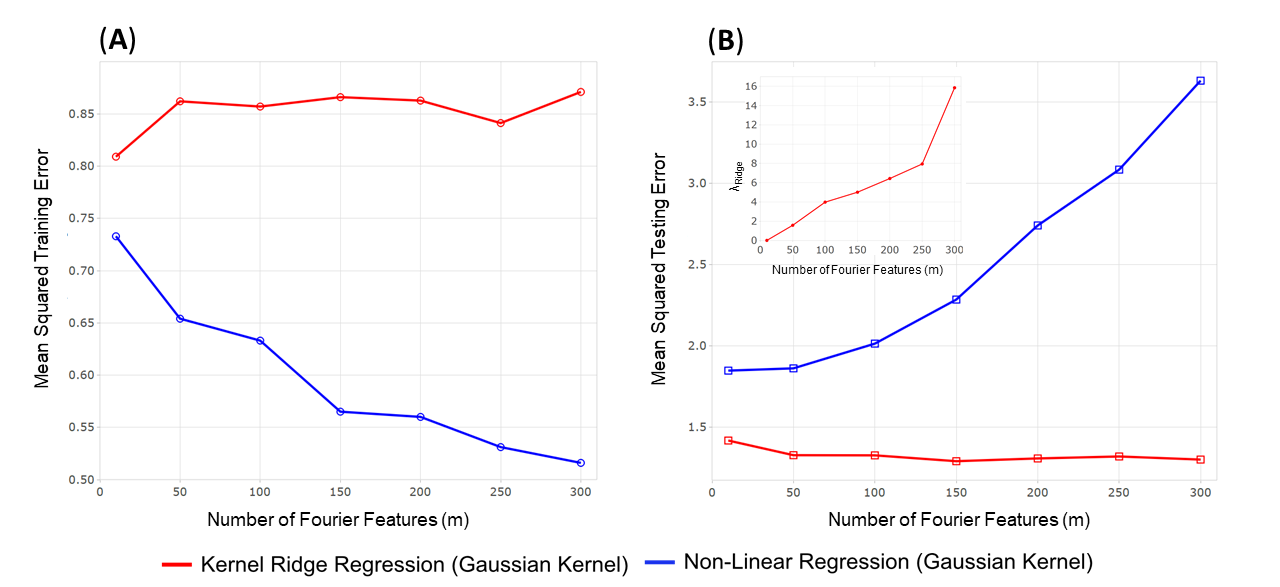}
  \caption{An example of the bias-variance trade-off for kernel regression (blue) and kernel ridge regression model (red) with a random Fourier feature approximation of a Gaussian kernel. The mean squared error (MSE) is calculated for both the (\textbf{A}) training data and the testing data (\textbf{B}) for an increasing number of sampled Fourier features, with optimal $\lambda_{Ridge}$ (by k-fold cross-validation) for the ridge regression model shown (\textbf{B} \textit{inset})}
  \label{fig.5}
\end{figure}

\section*{Advanced Methods for Random Fourier Features}
\subsection*{Limitations of RFF's}
Given the immense popularity and good empirical performance of the RFF method, little has been published on their limitations. However, RFF does have its limitations, including in the context of spatial analysis. From here onward, we term the RFF method described previously as the standard RFF method. 

Firstly, RFF's can be poor at capturing very fine scale variation as noted in Ton et al. \cite{TON201859}. This is likely due to fine-scale features being captured by the tails of the spectral density that will be infrequently sampled in the Monte Carlo integration.  Secondly, from a computational perspective, RFF's are very efficient but can still be poor compared to some state-of-the-art spatial statistics approaches. For example, the sparse matrix approaches based on SPDE as solution GMRF provide impressive savings with complexity $\mathcal{O}(m^{1.5})$ (compared to  $\mathcal{O}(m^{3})$ from the RFF primal solution) \cite{Lindgren}. Other methods such as the multiresolution Kernel approximation (MRA) provide incredible performance \cite{ding2017multiresolution}. However, it should be noted that many of these methods, such as MRA, are only valid for two dimensions \cite{ding2017multiresolution}, unlike RFF's which naturally extend to high-dimensions. Thirdly, while the convergence properties of RFF suggest excellent predictive capability \cite{rudi2017generalization}, alternative \emph{data-dependent} methods such as the Nystr{\"o}m approximation can perform much better in many settings \cite{Yang2012, rudi2015less}. The following sections will discuss the current methods that address some of these limitations.

\subsection*{Quasi-Monte-Carlo Features (QMC RFF)} 
One of the most significant limitations of standard RFF is the Monte Carlo integration. The infinite integral that describes the kernel function is converted to a finite approximation by sampling $\omega$'s from the spectral density. The convergence of Monte Carlo integration to the true integral occurs at the rate $\mathcal{O}(m^{0.5})$, which means for some problems a large number of features are required to approximate the integral accurately. 

A popular alternative is to use Quasi-Monte Carlo (QMC) integration \cite{Avrom2018}. In QMC integration, the set of points chosen to approximate the integral is chosen using a deterministic, low-discrepancy sequence \cite{niederreiter1978}. In this context, low-discrepancy means the points generated appear random even though they are generated from a deterministic, non-random process. For example, the Halton sequence,  that generates points on a uniform hypercube before transforming the points through a quantile function (inverse cumulative distribution function) \cite{Halton1964}. Low-discrepancy sequences prevent clustering and enforce more uniformity in the sampled frequencies, allowing QMC to converge at close to $\mathcal{O}(m^{-1})$ \cite{asmussen2007stochastic}. QMC can provide substantial improvements in the accuracy of the approximation of the kernel matrix for the same computational complexity. Crucially, QMC is trivial to implement within the RFF framework for some distributions. For example, for the squared exponential kernel, instead of generating features as by taking random samples from a Gaussian,  we generate them as: \newline

\begin{lstlisting}[language=R, caption=Example of Quasi-Monte Carlo sampling of a Gaussian power spectral density using a Halton sequence]
library(randtoolbox) #Package to generate Halton sequence
Omega = matrix(qnorm((halton(m,ncol(X)))),m) 
\end{lstlisting}

The remaining code for generating the features is exactly the same as Code 1. 

\subsection*{Leverage Score Sampling}
In the standard RFF method, frequencies are sampled with a probability proportional to their spectral density. However, the formula for the power spectral density, and concurrently sampling probability of a given frequency, is data-independent such that the formula for spectral density does not require data (see Table 1). Data-independent sampling is sub-optimal and can yield very poor results \cite{BachLS, mahoney2009cur, gittens2016revisiting}, and has been identified as one of the reasons RFFs perform poorly in certain situations \cite{li2018unified}. An alternative is a \emph{data-dependent}  approach, that considers the importance of various features \emph{given} some data. Several data-dependent approaches for RFF have been proposed \cite{Ionescu17, rudi2017generalization, li2018unified}, but one of the most promising and easiest to implement is sampling from the leverage distribution of the RFF (abbreviated to LRFF) \cite{li2018unified}.

Leverage scores are popular across statistics and are a key tool for regression diagnostics and outlier detection \cite{Hoaglin, Velleman}. The leverage score measures the importance a given observation will have on the solution of a regression problem. However, this perspective of leverage scores as a measure of importance can be extended to any matrix. The leverage scores of a matrix $A$ is given by the diagonal matrix $T = A(AA^{T})^{-1}A^{T}$. The leverage score for $i$-th row of matrix $A$ is equal to the $i$-th diagonal element in matrix $T$, denoted as $\tau_{i,i}$ and  calculated by:
\begin{equation}
\tau_{i,i} = \mathbf{a}_{i}^{T}(AA^{T})^{-1}\mathbf{a}_{i} = [A(AA^{T})^{-1}A^{T}]_{ii}
\end{equation}

$\tau_{i,i}$ can also be seen as a measure of the importance of the row $\textbf{a}_{i}$.

Most leverage score sampling methods apply ridge regularisation to the leverage scores,c ontrolled by regularisation parameter $\lambda_{\tiny{\textup{LRFF}}}$ given by:
\begin{equation}
\tau_{i,i}(\lambda) = \mathbf{a}_{i}^{T}(AA^{T}+ \lambda I)^{-1}\mathbf{a}_{i}
\end{equation}
The resulting scores are termed ridge leverage scores \cite{alaoui2014fast}. The regularisation serves a nearly identical purpose as when applied in the context of linear regression; ensuring the inversion required to compute the scores is always unique and less sensitive to perturbations in the underlying matrix, such as when only partial information about the matrix is known. The ridge parameter is key in stabilising leverage scores and permits fast leverage score sapling methods that approximate leverage scores using subsets of the full data \cite{musco2017recursive, rudi2018fast, drineas2012fast, cohen2017input}.

Leverage score sampling aims to improve the RFF method by sampling features with probability proportional to importance (rather than their spectral density). This is achieved by sampling columns of the feature matrix with a probability proportional to their leverage scores.  The resulting samples should contain more of the important features, and thus a more accurate approximation to the full matrix given the same number of samples. The computation and inversion of the matrix M increases the computational burden of this approach, but only has to be calculated once for a given regularization parameter. A key distinction to note is that the formula for the leverage score up to this point (and in the majority of the literature) have sought the leverage score of the rows. For RFF we want the leverage scores of the columns as they correspond to the Fourier bases. Therefore, the ridge leverage score of the Fourier features matrix is given by:
 
\begin{equation}
\begin{split}
    T(\lambda) & =diag\left(\Phi(X)(\Phi(X)^{T}\Phi(X) + \lambda_{\tiny{\textup{LRFF}}} I)^{-1}\Phi(X)^{T}\right) \\
    \tau_{i,i}(\lambda) & = \phi(\mathbf{x}_{i})(\Phi(X)^{T}\Phi(X) + \lambda_{\tiny{\textup{LRFF}}} I)^{-1}\phi(\mathbf{x}_{i})^{T}
\end{split}
\end{equation}

With suitable scaling, we can now sample Fourier features with a probability proportional to the leverage distribution, allowing us to sample features proportional to their importance. The code is given by: \newline

\begin{lstlisting}[language=R, caption=LRFF Example]
# Generate the m frequency samples (Omega) from the PSD
Proj = x %*% t(Omega)   # Projection
Phi = cbind(cos(Proj), sin(Proj)) / sqrt(m) # Basis
M = t(Phi) %*% Phi  # Primal matrix
T_lrff <- M %*% solve(M + n*diag(x=lambda,m)) 
pi_s <- diag(T_lrff) # Diagonal elements
l <- sum(diag(T_lrff))  #Sum of diagonal elements of T
is_wgt <- sqrt(pi_s/l)  # Leverage scores
\end{lstlisting}

\subsection*{Orthogonal Random Features}
One of the benefits of RFF's is that they can define kernels in high dimensions. For example, one can use a kernel in 4-dimensions to represent Cartesian spatial coordinates $x,y,z$ and time, $t$, the foundation for any spatiotemporal modelling. However, increasing dimensionality comes at a cost. While RFF's are unbiased estimators with respect to the expectation of the kernel matrix \cite{Rahimi2007}, increase the number of dimensions of the data significantly increases the variance of the RFF's estimate and requiring significantly more features to achieve an accurate approximation \cite{felix2016orthogonal}. One proposed approach to solve this issue with high dimensional kernels is to draw each new feature dimension orthogonally. 

In the standard RFF method, the sampled frequencies can be concatenated into a frequency matrix, $\Omega \in \mathbb{R}^{m \times d}$. If we consider a squared exponential/ Gaussian kernel, $\Omega$ is actually just a random Gaussian matrix, as the sampled frequencies are drawn from a standard normal distribution and scaled by the kernel parameter, $\sigma$. Therefore, the matrix $\Omega = \frac{1}{\sigma}G$, where $G$ is a random Gaussian matrix of dimension $\mathbb{R}^{d \times d}$ (and should not be confused with the gram matrix). 

In orthogonal random features (ORF), the aim is to impose orthogonality on $\Omega$, such that it contains significantly less redundancy than a random Gaussian matrix, capable of faster convergence to the full kernel matrix with lower variance estimates. The simplest method to impose orthogonality would be to replace $G$ with a random orthogonal matrix, $O$. However, if we consider the squared exponential/Gaussian kernel, the row norms of the $G$ matrix follow will Chi distribution. In comparison, the orthogonal matrix, $O$, will have (by definition) rows with unit norm. Thus, simply replacing $G$ with $O$ means that the RFF will no longer be an unbiased estimator. 

To return to an unbiased estimator, the orthogonal matrix, $O$, must be scaled by the diagonal matrix, $S$, with diagonal entries random variables from Chi distributed with $D$ degrees of freedom. This ensures that the row norms of $G$ and $SO$ will be identically distributed and can be used to construct a matrix of orthogonal random features given by $\Omega_{ORF} = \frac{1}{\sigma} SO$. It should be noted that the elements of $S$ are only Chi distributed random variables when $G$ is a random Gaussian matrix (the definition of a Chi distributed random variable identical to taking the L2 norm of a set of standard normally distributed variables). For other kernels, the diagonal elements of matrix $S$ are computed as the norm for the corresponding row in $G$. Therefore, the $i$-th diagonal element of $S$ is calculated by:
\begin{equation}
s_{i,i}=\left \| g_{i} \right \|_{2} = \sqrt{\sum_{j=1}^D \left (G_{i,j} \right )^2}    
\end{equation}

Therefore, generating orthogonal random features for a given kernel requires two steps. First, derive the orthogonal matrix $O$ by performing QR decomposition on the feature matrix $G$ (where $O$ corresponding to the Q matrix of QR decomposition). See \cite{gentle2012numerical} for an excellent summary of the QR decomposition. Second, compute the diagonal entries of the matrix, $S$, by talking the norms of corresponding rows of $G$. We then compute the orthogonal feature matrix as $\Omega_{ORF} = \frac{1}{\sigma} SO$. The random Fourier feature matrix is replaced with orthogonal random feature matrix to generate the orthogonal random basis matrix $\Phi_{ORF}(X)$, computed as $\Phi_{ORF}(X)  = [cos(X\Omega_{ORF}^{T}) \; sin(X\Omega_{ORF}^{T})] \in \mathbb{R}^{N \times 2m}$. The R code for ORFs is as follows: \newline

\begin{lstlisting}[language=R, caption=ORF example]
omega <- c()
while(Nrow < N_features){
	G <- matrix(rnorm(D1*D1),nrow=D1,ncol=D1) # Random Gaussian matrix
	G1 <- matrix(rnorm(D1*D1),nrow=D1,ncol=D1)
	O <- qr.Q(qr(G),complete=TRUE) #Orthogonal Matrix by QR decomposition
	S <- diag(sqrt(rowSums(G1^2))) #Diagonal Matrix
	omega1 <- rbind(omega,S  Q)
	Nrow <- nrow(omega)
} 
\end{lstlisting}

Yu et al. \cite{felix2016orthogonal} also included an extension to the ORF method termed structured ORF (SORF) to avoid the computationally expensive steps of deriving the orthogonal matrix ($\mathcal{O}(N^{3})$ time) and computing random basis matrix ($\mathcal{O}(N^{2})$ time). The SORF method replaces the random orthogonal matrix, $O$, by a class of specially structured matrices (consisting of products of binary diagonal matrices and Walsh-Hadamard matrices) that has orthogonality with near-Gaussian entries \cite{felix2016orthogonal}. The SORF method maintains a lower approximation error the standard RFF, but is significantly more computationally efficient than ORF, with computing $\Phi_{SORF}(X)$ taking only $\mathcal{O}(N log(N))$ time.

\subsection*{Non-stationary and Arbitrary Kernel Functions}
One of the most significant limitations to the standard RFF method is the restriction to shift-invariant kernels, where $K(x,z) = K(x-z)$. This restriction means that the kernel value is only dependent on the lag or distance $x-z$ rather than the actual locations. This property imposes stationarity on the spatiotemporal process. While this assumption is not unreasonable, and non-stationarity is often unidentifiable, in some cases the relaxation of stationary can significantly improve model performance \cite{paciorek2006spatial}.  

To extend the RFF method to non-stationary kernels requires a more general representation of Bochner's theorem capable of capturing the spectral characteristics of both stationary and non-stationary kernels. This extension \cite{yaglom2012correlation} states than any kernel (stationary or non-stationary) can be expressed as its Fourier transform in the form of:
\begin{equation}
    k(x_{1},x_{2}) = \int_{R^{d} \times R^{d}}e^{i(\omega^{T}_{1}x_{1}-\omega^{T}_{2}x_{2})} \mathbb{P}(\omega_{1}) \mathbb{P}(\omega_{2}) \; d\omega_{1} d\omega_{2}
\end{equation}

This equation is nearly identical to the original derivation of Bochner's theorem given in Equation 29, but now we have two spectral densities on $\mathbb{R}^{D}$ to integrate over. It is easy to how if the two spectral densities are the same, the function equates the definition for stationary kernels. Applying the same treatment and Monte Carlo integration can be performed to give the feature space of the non-stationary kernel \cite{TON201859}, now given by: 
\begin{equation}
    \Phi_{\tiny{\textup{RFF}}}(x)=
\begin{pmatrix}
 \cos(\omega^{1\;T} x) + \cos(\omega^{2\;T} x)\\
 \sin(\omega^{1\;T} x) + \sin(\omega^{2\;T} x)
\end{pmatrix}
\;\;\;\;\;\;\;, \left \{\omega^{\{1,2\}}\right\}_{j=1}^{m} \overset{i.i.d.} {\sim} \mathbb{P}^{\, \{1,2\}}(\omega)
\end{equation}
Note that this derivation requires drawing independent samples for both of the spectral densities, $\mathbb{P}^{\,l}(\omega)$, such that we generate two frequency matrices, $\Omega^{l} \in \mathbb{R}^{m \times d}$. 

In both the stationary and non-stationary case the choice of the kernel is often arbitrary or made with knowledge of the process being modelled. For example, if the spatial data is expected to be very smooth, then a squared exponential kernel can be used. It is, however, possible to treat  $\omega$ as unknown kernel parameters variables and infer their values \cite{TON201859}. This is equivalent to deriving an empirical spectral distribution. This strategy is data dependent and can achieve impressive results; however, great care must be taken to avoid overfitting \cite{TON201859}. 

\section*{Conclusion} 
Regression is a key technique for nearly all scientific disciplines and can be extended from its simplest forms to highly complex and flexible models capable of describing nearly any type of data. Within this paper, we have provided an overview of one such tool in random Fourier features. RFF allows for the extension of the linear regression model into a highly expressive non-linear regression model only using some trigonometric functions and can be implemented with only a few lines of code and provides significant computational benefits to working with a full kernel. To that end, random Fourier features and their extensions represent an exciting new tool for multi-dimensional spatial analysis on large datasets.

\section*{Supplementary Material}
\subsection*{Supplementary Equations 1 - Lagrangian Derivation of the Ridge Regression Dual Problem}
\setcounter{equation}{0}
Rather than solving the ridge regression objective function as a single minimisation problem, it can be considered as a constrained minimisation problem given. The solution of a constrained optimisation problem can often be found by using the so-called Lagrangian method. Given the constrained optimisation problem:
\begin{equation}
\begin{split}
    & \min f(x) \\
    s.t. \; \; \; & g(x) = r
\end{split}
\end{equation}

The constrained optimisation problem can be concatenated into a single equation termed the Lagrangian, given by:
\begin{equation}
    L(\mathbf{x},\mathbf{\alpha}, r) = f(\mathbf{x})+\alpha(g(\mathbf{x})-r) 
\end{equation}

The Lagrangian requires the introduction of a new variable, $\alpha$ called a Lagrange multiplier. The ridge regression can be converted into a constrained minimisation problem given by:
\begin{equation}
\begin{split}
    & \min_{\mathbf{w }\in \mathbb{R}^d} \left \|r\right \|^2 +\frac{\lambda}{2}\left\|\mathbf{w}\right \|^2 \\
    & s.t. \; \; \;  X\mathbf{w} - \mathbf{y} = r 
\end{split}
\end{equation}

with the Lagrangian given by:
\begin{equation}
    L(\mathbf{w},r,\alpha) = \frac{1}{2}\left \|r\right \|^2 +\frac{\lambda}{2}\left\|\mathbf{w}\right \|^2 + \alpha^{T}(X\mathbf{w} - \mathbf{y} - r)
\end{equation}

Within this context, $\alpha$,  are the dual variables. The aim set the derivatives with respect to the primal variables to zero and to solve for $\alpha$, that in this context represent the dual variables. The first step is to write the Lagrangian entirely in terms of the dual variables. Therefore, the first set is to find expressions of the primal variables $w$ and $r$ in terms of $\alpha$. This is achieved by setting the derivatives of the Lagrangian with respect to the primal variables to zero, giving:
\begin{equation}
\begin{split}
        \frac{\partial L(\mathbf{w},\alpha,r)}{\partial \mathbf{w}} & = \lambda \mathbf{w} - X^{T}\alpha, \; \; \; \; \mathbf{w} = \frac{1}{\lambda}X^{T}\alpha \\
         \frac{\partial L(\mathbf{w},\alpha,r)}{\partial r} & = r + \alpha, \; \; \; \; \; \; \; \; \; \; \; \; r = -\alpha
\end{split}
\end{equation}

Substituting the solutions for $\textbf{w}$ and $r$ back into the original Lagrangian equations gives the dual Lagrangian:
\begin{equation}
\begin{split}
     L(\mathbf{w}(\alpha),r(\alpha),\alpha) & = \frac{1}{2}\left \| \alpha\right \|^2 +\frac{\lambda}{2}\left\|X^{T}\alpha\right \|^2 + \alpha^{T}(\mathbf{y} - \frac{1}{\lambda}XX^{T}\alpha - \alpha) \\
       L(\mathbf{w}(\alpha),r(\alpha),\alpha) & = -\frac{1}{2}\left \| \alpha\right \|^2 -\frac{\lambda}{2}\left\|X^{T}\alpha\right \|^2 + \alpha^{T}\mathbf{y}
\end{split}
\end{equation}

Taking derivatives of the dual Lagrangian and setting it to zero allows the derivation of the optimal solution for the dual variables given by
\begin{equation}
    \mathbf{\alpha} = (XX^{T} + \lambda I_{n})^{-1}\mathbf{y}
\end{equation}

\subsection*{Supplementary Equations 2 - Feature Expansion of the Gaussian Kernel}
\setcounter{equation}{0}
The Gaussian kernel is a shift-invariant kernel given by:
\begin{equation}
   K(\mathbf{x},\mathbf{z}) = e \left (-\gamma {\left \| \mathbf{x}-\mathbf{z} \right \|^{2}} \right ), \; \; \; \; \gamma = \frac{1}{2\sigma^{2}}
\end{equation}

To examine the feature space associated with the kernel, let $\mathbf{x}$ and $\mathbf{z}$ be vectors in the space $R^{1}$ and $\gamma > 0$. The feature space can written as:
\begin{equation}
\begin{split}
    e^{-\gamma {\left \| \mathbf{x}-\mathbf{z} \right \|^{2}}} & = e^{ -\gamma (\mathbf{x}-\mathbf{z})^{2}} \\
    & = e^{-\gamma\mathbf{x}^{2} + 2\gamma\mathbf{xz} -\gamma\mathbf{z}^{2}} \\
    & = e^{(-\gamma\mathbf{x}^{2}-\gamma\mathbf{z}^{2})}e^{(2\gamma\mathbf{xz})} 
\end{split}
\end{equation}

We can apply a Taylor expansion to the second term:
\begin{equation}
    e^{(2\gamma\mathbf{xz})}= \left (1 + \frac{2\gamma \mathbf{xz}}{1!} + \frac{(2\gamma \mathbf{xz})^{2}}{2!} + \frac{(2\gamma \mathbf{xz})^{3}}{3!} + ...  \right ) = \sum_{n=0}^{\infty}\frac{(2\gamma \mathbf{xz})^{n}}{n!}
\end{equation}
where 
\begin{equation}
    K(\mathbf{x},\mathbf{z}) = e^{(-\gamma\mathbf{x}^{2}-\gamma \mathbf{z}^{2})}\sum_{n=0}^{\infty}\frac{(2\gamma \mathbf{xz})^{n}}{n!} = \left \langle \phi (\mathbf{x}),\phi (\mathbf{z}) \right \rangle
\end{equation}

Therefore, the feature mapping of vector $\textbf{x}$ is infinite, given by:
\begin{equation}
    \phi(\mathbf{x})=  e^{(-\gamma\mathbf{x}^{2})}\sum_{n=0}^{\infty}\sqrt{\frac{(2\gamma \mathbf{x})^{n}}{n!}}
\end{equation}
It should be noted that similar Taylor expansions can be applied to many kernels that contain exponential functions, such as Mat{\'e}n and Laplacian kernels, that also result in infinite feature spaces. 
\subsection*{Supplementary Equations 3 - Extended Derivation of Random Fourier Features form Bochner's Theorem}
\setcounter{equation}{0}

Bochner's theorem guarantees that an appropriately scaled shift-invariant kernel , where $k(x_{1},x_{2})=k(x_{1}-x_{2})$, is the inverse Fourier transform of a probability measure given by:
\begin{equation}
    k(x_{1}-x_{2}) = \int_{R^{d}}e^{i\omega^{T}(x_{1}-x_{2})} \mathbb{P}(\omega ) \; d\omega
\end{equation}
We can write this expectation as:
\begin{equation}
k(x_{1}-x_{2}) = \mathbb{E}_{\omega\sim \mathbb{P}}[e^{i\omega^{T}(x_{1}-x_{2})}]
\end{equation}

Applying Euler's identity to the exponential ($e^{i\pi}  = \cos (\pi) + i\sin (\pi)$), 
\begin{equation}
    \begin{split}
    k(x_{1}-x_{2}) & = \mathbb{E}_{\omega\sim \mathbb{P}}[e^{i\omega^{T}(x_{1}-x_{2})}] \\
    & = \mathbb{E}_{\omega\sim \mathbb{P}}[\cos(\omega^{T}(x_{1}-x_{2})) + i\sin (\omega^{T}(x_{1}-x_{2}))] \\
    \end{split}
\end{equation}

For the purpose of regression we only need to consider the real component:
\begin{equation}
    \begin{split}
    & = \mathbb{E}_{\omega\sim \mathbb{P}}[\cos(\omega^{T}(x_{1}-x_{2}))] \\
    & = \mathbb{E}_{\omega\sim \mathbb{P}}[\cos(\omega^{T}x_{1})\cos(\omega^{T}x_{2}) + \sin(\omega^{T}x_{1})\sin(\omega^{T}x_{2})] \\
    \end{split}
\end{equation}

By performing standard Monte Carlo integration on the real component, we can derive a finite-dimensional approximation of the kernel function as the following:
\begin{equation}
    \begin{split}
    & = \mathbb{E}_{\omega\sim \mathbb{P}}[\cos(\omega^{T}x_{1})\cos(\omega^{T}x_{2}) + \sin(\omega^{T}x_{1})\sin(\omega^{T}x_{2})] \\
    & \approx \sum_{j=1}^{m}\cos(\omega^{T}_{j}x_{1})\cos(\omega^{T}_{j}x_{2}) + \sin(\omega^{T}_{j}x_{1})\sin(\omega^{T}_{j}x_{2}) \;\;\;\;\;\;\;,\left \{\omega\right\}_{j=1}^{m} \overset{i.i.d}{\sim } \mathbb{P}(\omega)
    \end{split}
\end{equation}

By separating the $x_{1}$ and $x_{2}$ components the approximation to the kernel becomes:
\begin{equation}
    k(x_{1}-x_{2}) = \frac{1}{m}\sum_{j=1}^{m} 
\begin{pmatrix}
 \cos(\omega_j^T x_{1})\\
 \sin(\omega_j^T x_{1})
\end{pmatrix}^T
\begin{pmatrix}
 \cos(\omega_j^T x_{2})\\
 \sin(\omega_j^T x_{2})
\end{pmatrix}
= \Phi_{\tiny{\textup{RFF}}}(x_{1})\Phi_{\tiny{\textup{RFF}}}(x_{2})^T 
\;\;\;\;\;\;\;,\left \{\omega\right\}_{j=1}^{m} \overset{i.i.d}{\sim } \mathbb{P}(\omega)
\end{equation}

Thus, the new random Fourier feature matrix given by:
\begin{equation}
    \Phi_{\tiny{\textup{RFF}}}(x_{1}) = 
\begin{pmatrix}
 \cos(\omega_j^T x_{1})\\
 \sin(\omega_j^T x_{1})
\end{pmatrix} \;\;\;\;\;\;\;,\left \{\omega\right\}_{j=1}^{m} \overset{i.i.d}{\sim } \mathbb{P}(\omega) 
\end{equation}
with the equivalent mapping for $x_{2}$ shown by replacing the $x_{1}$ in the above equation with $x_{2}$.

\printbibliography

\end{document}